\documentclass{article}
\usepackage{spconf,amsmath,graphicx,diagbox}

\title{Acoustics based Intent Recognition using Discovered Phonetic Units for Low Resource Languages}
\name{Akshat Gupta, Xinjian Li, Sai Krishna Rallabandi, Alan W Black \thanks{akshatgu@andrew.cmu.edu, \{xinjianl, srallaba, awb\}@cs.cmu.edu}}
\address{Carnegie Mellon University}
\begin{document}

\maketitle


\begin{abstract}
With recent advancements in language technologies, humans are now speaking to devices. Increasing the reach of spoken language technologies requires building systems in local languages. A major bottleneck here are the underlying data-intensive parts that make up such systems, including automatic speech recognition (ASR) systems that require large amounts of labelled data. With the aim of aiding development of spoken dialog systems in low resourced languages, we propose a novel acoustics based intent recognition system that uses discovered phonetic units for intent classification. The system is made up of two blocks - the first block is a universal phone recognition system that generates a transcript of discovered phonetic units for the input audio, and the second block performs intent classification from the generated phonetic transcripts. We propose a CNN+LSTM based architecture and present results for two languages families - Indic languages and Romance languages, for two different intent recognition tasks. We also perform multilingual training of our intent classifier and show improved cross-lingual transfer and zero-shot performance on an unknown language within the same language family.
\end{abstract}

\section{Introduction}
\label{sec:intro}


In order to bring technology closer to humans, it is important to provide accessible mechanisms of interaction. Speech is regarded as the most natural form of interaction for humans and its accessibility has been aided by improvements in language technologies such as Automatic Speech Recognition and Speech Synthesis. However, lack of annotated data is a major bottleneck for scaling speech technologies to new languages and domains. Therefore, it is useful to design techniques that can perform well in low data scenarios. A fundamental resource required to build such systems is a phonetic lexicon which can translate acoustic input to textual representation. In this paper, we present a novel approach to perform intent recognition purely from acoustics using such a phonetic lexicon. A block diagram representing our system is shown in Figure \ref{fig:sys-block}. Through our approach, we bypass the need to build language specific ASR systems, which are very data intensive, and demonstrate deployable performance for intent recognition using discovered phonetic units.



We test the performance of our system on two language families - Indic and Romance languages, each having a different intent recognition task. We also train our intent classification system multilingually and evaluate its zero-shot performance for a language not in the training set, although still within the same language family. This simulates a zero resource scenario and helps us understand the extent of cross-lingual transfer between languages for our acoustics based intent recognition system. We find that a multilingually trained classification model performs significantly better than a monolingually trained model for an unknown language. 

\section{Related Works}
\label{sec:related}



Spoken Language Understanding (SLU) systems aim to process spoken utterances for various downstream tasks. Current research in high resourced languages is moving towards building end-to-end SLU systems \cite{lugosch2019speech} \cite{radfar2020end} to eliminate propagating errors through the SLU pipeline. A typical SLU pipeline is made up of two blocks - a speech to text module followed by a natural language understanding (NLU) module. Building speech to text modules require large amounts of labelled speech data, which is scarce for low resourced languages. \cite{karunanayake2019transfer} \cite{karunanayake2019sinhala} have previously used outputs of an English ASR system and English phonemes for intent classification in Sinhala and Tamil. 

To the best of our knowledge, this paper is the first attempt in literature to use the phonetic units for intent classification. There have been numerous attempts\cite{badino_autoencoder, hmm-vae_bhiksha} to discover such acoustic units in an unsupervised fashion. In \cite{subword_diarization}, authors presented an approach to modify the speaker diarization system to detect speaker-dependent acoustic units. \cite{unsupervised_AMtraining_ArenJansen} proposed a GMM-based approach to discover speaker-independent subword units. However, their system requires a separate Spoken Term Detector. Our work is closest to \cite{hillis2019unsupervised} where authors discover symbolic units in an unsupervised fashion for speech to speech translation.  Contrary to this work, we employ the symbolic units generated by Allosaurus \cite{li2020universal} which is trained in a supervised fashion. 


\section{DATASETS}
\label{sec:Datasets}

We study the performance of our acoustics based intent recognition system for two language families - Indic Languages and Romance Languages. For each family we use a different dataset and each language family has a different intent recognition task.

\subsection{Dataset for Indic Languages}
\label{ssec:data_indic}
We use Google's Taskmaster-1 Dataset \cite{byrne2019taskmaster} for Indic Languages which contains data for user interactions with an autonomous dialogue system collected using the Wizard of Oz methodology \cite{hanington2012universal}. The user dialogues are a written transcripts of the conversations in English. The dataset contains labelled intents and slots for the conversation. We extract the sentence responsible for the labelled intent from the dataset and create an intent recognition dataset. We obtain 3243 utterances in total distributed amongst 6 intents as shown in Table \ref{Table1}. 
\begin{table}
\centering
\begin{tabular}{cc}
\hline
\multicolumn{1}{|p{3cm}|}{\centering \textbf{Intents}} &  \multicolumn{1}{|p{3.5cm}|}{\centering \textbf{Number of Utterances }}\\
\hline
ordering pizza & 711 \\
auto-repair appointment  & 484\\
order ride service & 450 \\
order movie tickets & 549\\
order coffee & 292\\
restaurant reservations & 757\\
\hline
\end{tabular}
\caption{Class distribution for the Indic dataset. }\label{Table1}
\end{table}
After creating the intent classification dataset, we translate the transcripts in Engish into four Indic languages - Hindi, Gujarati, Bengali and Marathi, using the Google Translate API. The translated text was used to synthesize audio using the Google Text-To-Speech API for Hindi, Gujarati and Bengali. CLUSTERGEN \cite{black2015random} was used synthesizing for Marathi voice. The voice quality of Marathi is much worse when compared to the other voices generated from Google's API. The dataset in each language contains two voices - one male and one female. The audios are then passes into Allosaurus \cite{li2020universal} to discover phonetic units and create a phonetic transcription of the audio. 

\subsection{Dataset for Romance Languages}
To work with Romance languages, we create an intent recognition dataset from the MultiWoz dataset \cite{budzianowski2018multiwoz}. The dataset contains a large number of dialogues between humans and robots where each utterance is associated with a json object containing the conversational context. The context has rich information about the intent of humans. The largest context class in the dataset is about the reservation day whose distributions are shown in the Table \ref{Table2}. This class is used as our Romance languages dataset. This dataset is then prepared in a similar way as done for the Indic language, where we translate the original English utterances into 4 different Romance languages - Italian, Portuguese, Romanian and Spanish. The translated text is synthesized with the Google TTS engine, and then transcribed into phonetic units with Allosaurus \cite{li2020universal}.

\begin{table}
\centering
\begin{tabular}{cc}
\hline
\multicolumn{1}{|p{3cm}|}{\centering \textbf{Intents}} &  \multicolumn{1}{|p{3.5cm}|}{\centering \textbf{Number of Utterances }}\\
\hline
Monday & 743 \\
Tuesday  & 718 \\
Wednesday & 757 \\
Thursday & 738\\
Friday & 763\\
Saturday & 779\\
Sunday & 806\\
\hline
\end{tabular}
\caption{Class distribution for the Romance dataset.}\label{Table2}
\end{table}

\section{MODELS}
\label{sec:models}
\begin{figure}[htb]

\begin{minipage}[b]{1.0\linewidth}
  \centering
  \centerline{\includegraphics[width=9.5cm]{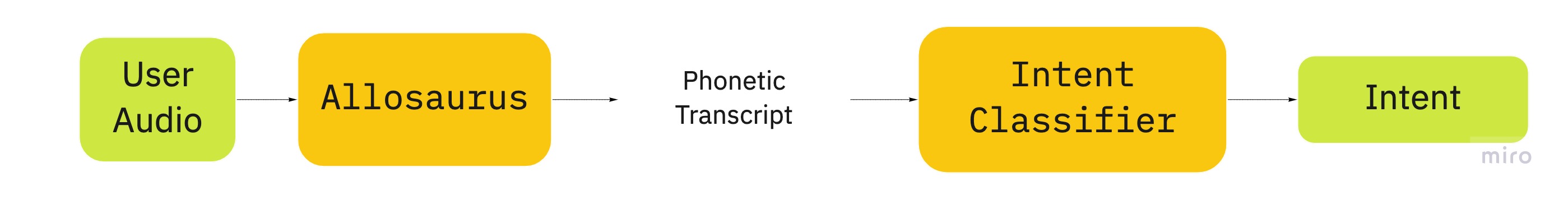}}
\end{minipage}
\caption{Block Diagram showing a general acoustics based intent recognition system.}
\label{fig:sys-block}
\end{figure}

A block diagram depicting our acoustics based intent recognition system utilizing a phonetic transcription is shown in Figure \ref{fig:sys-block}. The input audio is directly fed into a system that can generate hypothesized phonetic units. For our work, we use the Allosaurus library \cite{li2020universal} which is a nearly-universal phone recognition system. For this work, we employ the language dependent phones, which basically means we're providing an identifier to Allosaurus for audio language. The phonetic transcription is then sent to an intent classifier that does the classification purely based on the generated sequence of phones. Such very simple systems can be used to build powerful tools, especially for low resource languages, as shown in \cite{gupta2020mere}.

\begin{figure}[t]

\begin{minipage}[b]{1.0\linewidth}
  \centering
  \centerline{\includegraphics[width=6cm]{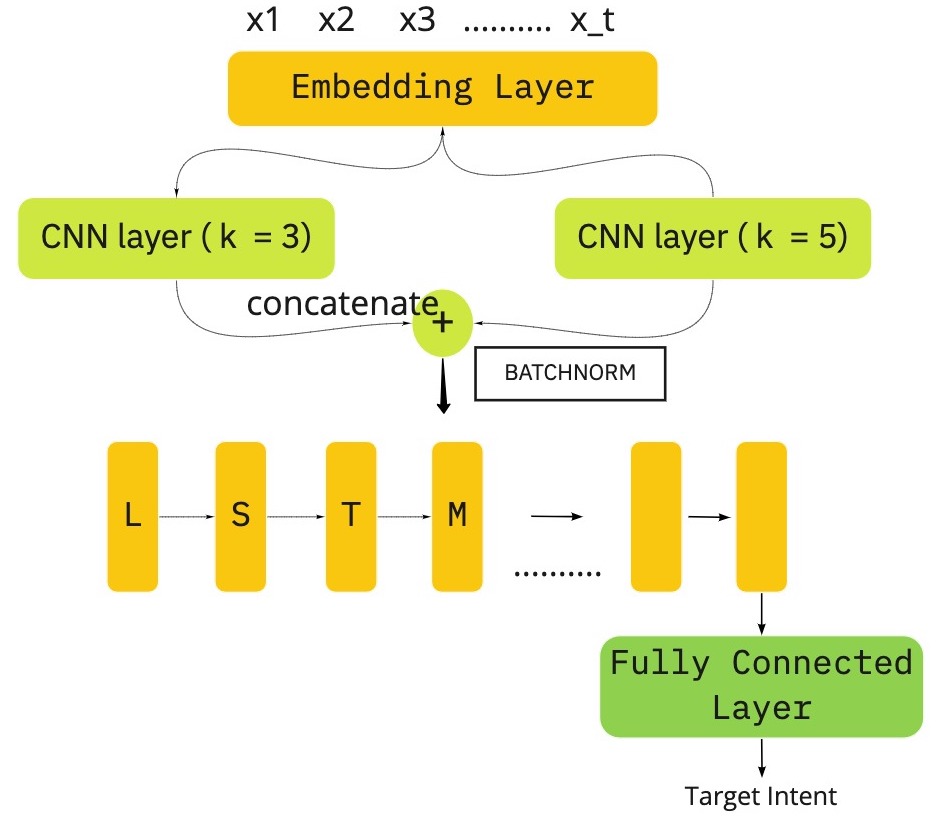}}
\end{minipage}
\caption{Block Diagram depicting the architecture of our proposed neural network.}
\label{fig:arch}
\end{figure}

We use a Naive Bayes classifier as our baseline with add-1 smoothing and absolute discounting. We also propose a neural network architecture shown in Figure \ref{fig:arch} to compare with the baseline results. The architecture is based on LSTMs (long-short term memory) \cite{hochreiter1997long} for modeling sequential information where the contextual information is encoded using CNNs (convolutional neural network).  

The input to the network is a sequence of phones $\mathbf{x} = {x_1, x_2, \dots x_t}$, where each phonetic unit is passed to a 128 dimensional embedding vector. The embedding layer converts the input sequence into a dense vector representation which is then sent to two 1-d CNN layers of kernel size k = 3,5. The CNN layers have 128 filters and capture trigram and 5-gram features from the phonetic transcription. The outputs of each of the CNN layers are concatenated to create a 256 dimension long embedding vector where each embedding vector now has contextual information encoded in it.

The concatenated embeddings are passed through the LSTM layer consisting of 128 neurons. The hidden state of the LSTM layer at the final time step is sent to a linear layer for intent classification. 

\section{EXPERIMENTS}
\label{sec:experiments}
We test our acoustics based intent recognition system for two sets of languages across two different language families - Indic and Romance language families. We perform monolingual and multilingual training for both baseline and our proposed neural network architecture and test the model performance for multiple languages. 

\subsection{Monolingual Training Results}
\label{ssec:monolingual}
In this section we present results for intent classification architectures trained on a single language. Table \ref{Table3} presents the classification results for Indic languages and Table \ref{Table4} for Romance languages. The diagonal elements in the tables show the classification accuracy for training and testing performed on the same language. The numbers in the bracket show performance with the baseline (Naive Bayes) classifier. We see that our proposed neural network architecture improves on our baseline significantly.

Cross-lingual testing results for monolingually trained classification models are also shown in Tables \ref{Table3} and \ref{Table4}. The performance is relatively poor when the classification model is trained on only one language due to minimal cross-lingual transfer. Language pairs for linguistically similar languages show higher performance. This can be seen for language pairs Hindi-Gujarati and Gujarati-Marathi in Indic language family and pairs Italian-Portuguese and Italian-Spanish in the Romance language family. These language pairs are also geographically close. The cross lingual results are in general better for the Indic Dataset when compared to the Romance dataset. We believe this is because all Indic languages have some amount of code mixing within them. Therefore, there is a larger cross-lingual transfer of features between any pair of languages in the Indic language family when compared to the Romance language. 

\begin{table}
\centering
\begin{tabular}{|c|cccc|}
\hline
\multicolumn{1}{|p{1cm}|}{ \textbf{\backslashbox[1.4cm]{Train}{Test}}} & \multicolumn{1}{|p{1cm}|}{ \textbf{Hin}} & 
\multicolumn{1}{|p{1cm}|}{ \textbf{Guj}} & 
\multicolumn{1}{|p{1cm}|}{\textbf{Mar}} & 
\multicolumn{1}{|p{1cm}|}{\textbf{Ben}}\\
\hline
\textbf{Hin} & \textbf{92.0}(89.3) & 54.7(59.7) & 43.7(36.7) & 54.3(45.3)\\
\textbf{Guj} & 52.3(50.3) & \textbf{93.3}(91.7) & 52.0(47.0) & 63.0(39.3)\\
\textbf{Mar} &  52.0(35.0) & 66.3(49.7) & \textbf{87.7}(84.3) & 58.0(37.0)\\
\textbf{Ben} & 48.0(41.7) & 54.7(38.3) & 45.7(31.3) & \textbf{95.0}(93.0)\\
\hline
\end{tabular}
\caption{Classification Accuracy for monolingual training for Indic Languages - Hindi (Hin), Gujarati (Guj), Marathi (Mar) and Bengali (Ben). The numbers in the bracket are the baseline results using a Naive Bayes classifier.}\label{Table3}
\end{table}

\begin{table}
\centering
\begin{tabular}{|c|cccc|}
\hline
\multicolumn{1}{|p{0.8cm}|}{\textbf{\backslashbox[1.3cm]{Train}{Test}}} & \multicolumn{1}{|p{1cm}|}{\textbf{Hin}} & 
\multicolumn{1}{|p{1cm}|}{\textbf{Guj}} & 
\multicolumn{1}{|p{1cm}|}{\textbf{Mar}} & 
\multicolumn{1}{|p{1cm}|}{\textbf{Ben}}\\
\hline
\textbf{HGM} & 85.3(84.7) & 90.3(86) & 75.6(78.3) & \textbf{80.7(58.3)}\\
\textbf{HGB} & 87.3(84) & 90.0(84) & \textbf{61.7(54.3)} & 90.3(89.3)\\
\textbf{HMB} & 84.3(88.7) & \textbf{62(65.4)} & 80.7(76.6) & 88.3(88)\\
\textbf{GMB} & \textbf{65.3(63.7)} & 86.7(84.7) & 83.0(80) & 92.0(89.7)\\
\hline
\end{tabular}
\caption{Average Classification Accuracy for a multilingually trained model. The languages in bold are the languages that are not present in the train set. The numbers in the bracket are the baseline results using a Naive Bayes classifier.}\label{Table5}
\end{table}

\subsection{Multilingual Training Results}
\label{ssec:multingual}
With the aim of improving performance on a language not present in our training set and simulating a zero resource scenario, we train a multilingual model. The training set size is kept the same and the exact same train-test split is used for accuracy scores as used for monolingual results. Let T = [$L_1$, $L_2$, \dots $L_n$] be the set of languages we use to train the classifier. Then the training set is divided randomly and equally amongst the '$n$' languages present in the training set. 

The results for multilingual training can be seen in Table \ref{Table5} and Table \ref{Table6} when trained on $n=3$ languages. The results in bold are for the language not in the training set. The numbers in the bracket show performance with the baseline (Naive Bayes) classifier. We see that our proposed neural network architecture improves on the baseline results significantly in almost all cases. The power of multilingual training becomes apparent when we look at the performance on a language not present in T. We find that a multilingual classifier always performs better on  an unknown language $L_u \notin T$ when compared to zero-shot transfer by monolingual model without significant performance loss in individual languages. Its important to consider that we haven't augmented the data in any form, thus the multilingual model see far fewer example of a specific language than monolingual models. 

The results in Table \ref{Table5} and Table \ref{Table6} show that there is larger amount of cross-lingual transfer when the model is trained on many languages from the same language family. In a practical scenario, this means that a deployed multilingual model is more likely to generalize better to an unknown language or variation in dialect than a monolingual model. This is especially useful for the case of low-resource languages for which it's hard to collect any training data. We illustrate this point by taking the example of Bengali in the Indic language family. There are four models across Tables \ref{Table3} and \ref{Table5} that do not have Bengali in the training set. These are the models not highlighted in the Bengali column in Table \ref{Table3} and the highlight value in the Bengali column in Table \ref{Table5}. The multilingual model performs best out of these four models. This is also true for Spanish in Tables \ref{Table4} and \ref{Table6}.

\begin{table}
\centering
\begin{tabular}{|c|cccc|}
\hline
\multicolumn{1}{|p{1cm}|}{ \textbf{\backslashbox[1.4cm]{Train}{Test}}} & \multicolumn{1}{|p{1cm}|}{ \textbf{Ita}} & 
\multicolumn{1}{|p{1cm}|}{ \textbf{Por}} & 
\multicolumn{1}{|p{1cm}|}{\textbf{Ron}} & 
\multicolumn{1}{|p{1cm}|}{\textbf{Spa}}\\
\hline
\textbf{Ita} & \textbf{88.6}(82.4) & 27.6(24.2) & 28.6(33.9) & 39.0(32.5)\\
\textbf{Por} & 30.6(23.3) & \textbf{88.6}(74.6) & 28.2(22.2) & 36.0(26.2)\\
\textbf{Ron} & 32.6(31.6) & 29.6(17.4) & \textbf{86.6}(76.5) & 46.0(33.3)\\
\textbf{Spa} & 46.1(35.8) & 43.9(40.2) & 35.4(33.9) & \textbf{88.7}(83.3)\\
\hline
\end{tabular}
\caption{Classification Accuracy for monolingual training for Romance Languages - Italian (Ita), Portuguese (Por), Romanian (Ron and Spanish (Spa). The numbers in the bracket are the baseline results using a Naive Bayes classifier.}\label{Table4}
\end{table}

\begin{table}
\centering
\begin{tabular}{|c|cccc|}
\hline
\multicolumn{1}{|p{1cm}|}{ \textbf{\backslashbox[1.4cm]{Train}{Test}}} & \multicolumn{1}{|p{1cm}|}{ \textbf{Ita(I)}} & 
\multicolumn{1}{|p{1cm}|}{ \textbf{Por(P)}} & 
\multicolumn{1}{|p{1cm}|}{\textbf{Ron(R)}} & 
\multicolumn{1}{|p{1cm}|}{\textbf{Spa(S)}}\\
\hline
\textbf{IPR} & 87.0(78.8) & 88.9(62.1) & 85.4(69.1) & \textbf{60.4}(43.9)\\
\textbf{IPS} & 88.6(77.2) & 88.9(62.9) & \textbf{37.4}(40.7) & 88.9(80.4)\\
\textbf{IRS} & 88.1(88.2) & \textbf{41.3}(30.0) & 86.6(73.2) & 88.7(80.0) \\
\textbf{PRS} & \textbf{50.3}(40.9) & 87.8(59.9) & 84.9(69.1) & 88.3(79.9) \\
\hline
\end{tabular}
\caption{Classification Accuracy for a multilingually trained model. The languages in bold are the languages that are not present in the train set. The numbers in the bracket are the baseline results using a Naive Bayes classifier.}\label{Table6}
\end{table}

\begin{figure}[htb]

\begin{minipage}[b]{1.0\linewidth}
  \centering
  \centerline{\includegraphics[width=6.5cm]{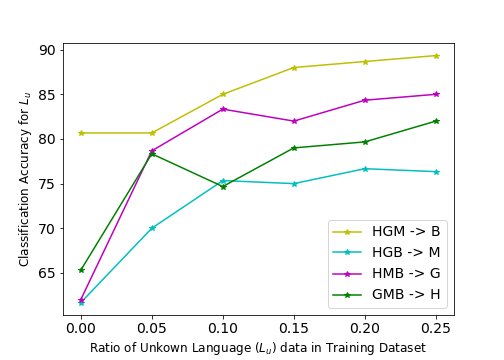}}
\end{minipage}
\caption{Plot showing performance of a multilingual intent classification model for when data for a language is injected into the training set in increments of ratio of 0.05 for Indic languages. For example, HGM -$>$ B represents a model trained on Hindi, Gujarati, Marathi and we're checking the increase in perfromance on Bengali by injecting Bengali data into the training dataset.}
\label{fig:variation}
\end{figure}

The performance for an unknown language $L_u \notin T$ can further be improved by injecting a very small amount of data for $L_u$ in the training set. We added training data for language $L_u$ in increments of a ratio of 0.05 of the training set as shown in Figure \ref{fig:variation}. We see that introducing even the slightest amount of training data for the unknown language increases its performance significantly while not affecting the performance of the other languages. Figure \ref{fig:variation} shows an increase in performance of about 9\% for Marathi, 14\% for Hindi and 17\% for Gujarati only by an injection of data 5\% the size of training dataset. 

\section{DISCUSSION}
\label{sec:discussion}
We present a novel approach for intent recognition in low resourced languages with experiments on two different language families. It was shown that zero-shot performance for a language not in the training set of the model but still within the language family can be improved with multilingual training. This helps in maximal cross-lingual transfer between languages that are linguistically and geographically closer to each other. We also found that performance for such a language not in the training set can be improved significantly by introducing a minimal amount of training data. 

Our present work was based on synthesized data due to the absence of enough natural speech datasets for intent recognition for low resource languages. Future work can include corroboration of our results with natural speech. The synthesized speech also had little speaker variation in terms of speaker style or prosody though we did include variation in speaker gender. 

\section{CONCLUSION}
\label{sec:conclusion}
We present a novel acoustics based intent recognition system that classifies intents from phonetic transcripts generated using a (nearly-)universal phone recognizer, bypassing the need to build language specific ASR. We also show that multilingual training within same language families produce better zero shot transfer within same the language family when compared to monolingual models.

\bibliographystyle{IEEEbib}
\bibliography{strings,refs}

\end{document}